\def\year{2017}\relax
\documentclass[letterpaper]{article} 
\usepackage{aaai17}  
\usepackage{times}  
\usepackage{helvet}  
\usepackage{courier}  
\usepackage{url}  
\usepackage{graphicx}  
\usepackage{graphics}
\usepackage{subcaption}
\usepackage{float}
\usepackage[noend,linesnumbered,ruled,vlined]{algorithm2e}
\usepackage{amsmath}
\usepackage{multirow}
\usepackage{longtable}
\usepackage{lscape}
\usepackage{lipsum}
\usepackage{fixltx2e}
\usepackage{bbm}

\makeatletter
\newcommand*{\centerfloat}{%
  \parindent \z@
  \leftskip \z@ \@plus 1fil \@minus \textwidth
  \rightskip\leftskip
  \parfillskip \z@skip}
\makeatother

\newcommand{\open}{\mbox{OPEN}\xspace}

\newcommand{\focal}{\mbox{FOCAL}\xspace}

\begin{document}
%
\title{Rapid Randomized Restarts for Multi-Agent Path Finding Solvers}
\author{Liron Cohen, Glenn Wagner, T.K. Satish Kumar, Howie Choset \and Sven Koenig \\ University of Southern California \\ Carnegie Mellon University}
\maketitle
\begin{abstract}
Multi-Agent Path Finding (MAPF) is an NP-hard problem well studied in artificial intelligence and robotics. It has many real-world applications for which existing MAPF solvers use various heuristics. However, these solvers are deterministic and perform poorly on ``hard'' instances typically characterized by many agents interfering with each other in a small region. In this paper, we enhance MAPF solvers with randomization and observe that they exhibit heavy-tailed distributions of runtimes on hard instances. This leads us to develop simple rapid randomized restart (RRR) strategies with the intuition that, given a hard instance, multiple short runs have a better chance of solving it compared to one long run. We validate this intuition through experiments and show that our RRR strategies indeed boost the performance of state-of-the-art MAPF solvers such as iECBS and M*.
\end{abstract}

\section{INTRODUCTION}
  Given an environment and agents with assigned start and goal locations, the Multi-Agent Path Finding (MAPF) problem is to find collision-free paths for all agents from their start to their goal locations that optimize some criterion such as makespan or sum of traveled distances.  Solving the MAPF problem has many applications, including improving traffic at intersections, search and rescue, formation control, warehouse applications, and assembly planning.  A comprehensive  list  of  applications  with  references  can  be found  in  \cite{L:BOOK:06}.

In artificial intelligence (AI), the MAPF problem is studied with the following simplifying assumptions. Given a directed or undirected graph and a set of agents with unique start and goal vertices, the MAPF problem is to find collision-free paths for all agents from their respective start vertices to their respective goal vertices. The agents traverse edges in unit time but can also wait at vertices. Even with these simplifying assumptions, minimizing the solution cost given by the makespan or sum of travel times of the agents along their paths is NP-hard \cite{YL:AAAI:13}.

Although the MAPF problem is NP-hard even with these simplifying assumptions, existing discrete AI MAPF solvers\footnote{henceforth simply referred to as `MAPF solvers'} perform well in practice in terms of runtime and solution cost. The simplifying assumptions made by these solvers is not debilitating for real-world applications because a polynomial-time post processing phase can reinstate kinematic constraints of the agents to make the solvers' plans executable on real robots \cite{HKCMXAK:ICAPS:16}.

Different techniques have been used to develop a variety of MAPF solvers. These include reductions to problems from satisfiability (SAT), integer linear programming and answer set programming.\footnote{The reader can find a list of references in \cite{MKACHKUXTS:IJCAI:16}.} However, current state-of-the-art MAPF solvers are based on graph search. These search-based solvers exploit available opportunities for decoupling agents when no coordination between them is required (for example, when they operate in spatially non-overlapping regions). Moreover, search-based solvers can also easily incorporate additional domain-specific constraints like downwash constraint that prevent one quadcopter from flying directly above another. To our knowledge, there are two families of search-based solvers that are state-of-the-art: M* \cite{W:DISS:15} and CBS \cite{SSFS:AIJ:15}.

M* is an A*-based algorithm that uses subdimensional expansion to initially create a one-dimensional search space embedded in the joint configuration space of the multi-agent system. When the search space is blocked because of an agent-agent collision, the dimensionality of the search space is locally increased to ensure that an alternative path can be found. Like M*, CBS also tries to avoid operating in the joint configuration space. However, it does so using a two-level search. In the high-level, a search is performed on a conflict tree that represents conflicts between individual agents. Each high-level node represents a set of constraints imposed on the motions of individual agents. In the low-level, single-agent searches are performed in conformance with the constraints imposed by the relevant high-level nodes. The basic versions of both M* and CBS generate optimal solutions.

Since the MAPF problem is NP-hard, even M* and CBS fail to scale for instances that have many agents interfering with each other in a small region. However, suboptimal versions of M* and CBS are capable of solving harder MAPF instances by trading off runtime with solution cost. Bounded-suboptimal MAPF solvers generate solutions with costs no more than a constant factor away from the optimal solution's cost. Bounded-suboptimal versions of M* and CBS use a user-specified constant for their suboptimality bounds.

As in weighted-A*, a $w$-suboptimal version of M* uses inflated heuristic values. Inflating heuristic values by a constant $w$ potentially speeds up planning because it focuses search to where solutions can be found. Enhanced CBS (ECBS($w$)) \cite{BSSF:SOCS:14} is a $w$-suboptimal variant of CBS. Its suboptimality comes from its use of focal search \cite{PJ:IEEE:82} with parameter $w$. Larger values of $w$ result in greedier searches. CBS with highways (CBS+HWY($w$)) \cite{CUK:SOCS:15} is also a $w$-suboptimal variant of CBS. However, its suboptimality comes from inflating heuristic values non-uniformly using highways. The idea of highways originates from Experience Graphs \cite{PCCL:RSS:12}. Highways are a set of ``useful'' directed edges in the graph that are either human-generated or automatically generated \cite{CUKXAK:IJCAI:2016}. Given a highway, the highway-aware heuristic inflates the costs of move actions along edges that do not belong to the highway by a factor of $w$. This biases the low-level searches to find paths that use the highway edges, reducing the number of head-on collisions. Highways can also be used in M* with similar inflation mechanisms and suboptimality guarantees. Improved ECBS($w$) (iECBS($w$)) \cite{CUKXAK:IJCAI:2016} is yet another $w$-suboptimal MAPF solver that successfully uses both highways and focal search.

Suboptimal MAPF solvers are capable of solving harder instances compared to optimal ones. However, even their runtimes increase exponentially with increased coupling among the agents. In part, the solvers' inability to cope with strongly coupled instances comes from the deterministic nature of their search processes. This weakness of deterministic search algorithms has also been observed in other combinatorial tasks, such as in SAT and constraint satisfaction problems (CSPs) \cite{GSCK:JAR:00}. Given a problem instance, deterministic search processes build the same search tree in every run. This means that ``bad'' decisions, especially higher up in the search tree, are not only expensive but also cannot be avoided in subsequent runs.

In this paper, we enhance search-based deterministic MAPF solvers with randomization. Some of the arbitrary, but deterministic, decisions that shape the search trees produced by these solvers are replaced with random ones. Given a MAPF instance, the runtime of such a randomized MAPF solver is a random variable. This random variable has been observed to exhibit a heavy-tailed distribution \cite{CUKXAK:IJCAI:2016}. Heavy-tailed distributions in runtimes can be exploited using a rapid randomized restart (RRR) strategy \cite{GSCK:JAR:00}. An RRR strategy generally uses the intuition that, given a hard instance, multiple short runs have a better chance of solving it compared to one long run.

In the experimental results section of this paper, we validate the effectiveness of RRR strategies for randomized MAPF solvers. We show that RRR strategies increase the success rate in CBS as well as M*-based approaches.

\section{BACKGROUND}
The MAPF problem is formally defined as follows: We are given a directed or undirected graph $G = (V,E)$ and a set of $K$ agents $1, \ldots, K$. Each agent $j$ has a unique start vertex $s^j \in V$ and a unique goal vertex $g^j \in V$. At each time step, each agent can either move to a neighboring vertex or wait at its current vertex, both with cost one. A \emph{solution} of a MAPF instance is a set of feasible paths, one path $\{ s^j_0, \ldots, s^j_{T_j}, s^j_{T_j +1}, \ldots \}$ for each agent $j \in \{1, \ldots, K\}$, such that no two paths collide. A path for agent $j$ is \emph{feasible} if and only if 1) it starts at the start vertex of agent $j$, that is, $s^j_0=s^j$; 2) it ends at the goal vertex of agent $j$ and remains there, that is, there exists a smallest $T_j$ such that $s^j_{T_j}=g^j$ and, for each $t > T_j$, $s^j_t = g^j$; and 3) every action is a legal move or wait action, that is, for all $t \in \{0,1, \ldots, T_j-1\} \ , \ \langle s^j_t,s^j_{t+1} \rangle \in E$ or $s^j_t = s^j_{t+1}$. A \emph{collision} between the paths of agents $j$ and $k$ is either a \emph{vertex collision} $(j,k,s,t)$, that is, $s = s^j_t = s^k_t$, or an \emph{edge collision} $(j,k,s_1,s_2,t)$, that is, $s_1 = s^j_t = s^k_{t+1}$ and $s_2 = s^j_{t+1} = s^k_t$. The travel time of agent $j$ is the number of time steps $T_j$ until it reaches its goal vertex. Our objective is to minimize the solution cost, given as the sum of the travel times of all agents, which is a common objective in the literature \cite{YL:AAAI:13,SSFS:AIJ:15}.

M* \cite{W:DISS:15} is an optimal MAPF solver that implements an approach called subdimensional expansion. It starts by computing an individual policy for each agent separately (that is, disregarding all other agents). An individual policy describes actions that move an agent along the optimal path to its goal. M* then uses the individual policies to define a search graph that is a small subgraph of the joint configuration graph, which is explored using A*.  Whenever an agent-agent collision is found, the search graph is grown to allow the colliding agents to consider alternative paths. M* defines the search graph by considering only a limited set of neighbors of each vertex, $v_k$, in the joint configuration space. By default, each agent follows its individual policy, producing a single limited neighbor; but this usually leads to agent-agent collisions. When M* finds a collision, it adds the colliding agents to the collision sets of all vertices that precede this collision in the search tree. The collision set, $C_k$, of $v_k$ therefore consists of all agents that may need to depart from their individual policies at $v_k$ to avoid a collision. The limited neighbors of $v_k$ are then generated by considering all possible actions for the agents in $C_k$ while restricting each agent not in $C_k$ to its individual policy. Whenever the collision set of a vertex is changed, the vertex is added back to the A* \open list, allowing M* to backtrack and consider alternate paths through previously expanded vertices. Inflated-M* is a bounded-suboptimal version of M* that uses inflated heuristic values similar to weighted-A* \cite{P:AI:70}.

CBS \cite{SSFS:AIJ:15} is an optimal MAPF solver. It performs high-level and low-level searches. Each high-level node contains a set of constraints and, for each agent, a feasible path that respects the constraints. The high-level root node has no constraints. The high-level search of CBS is a best-first search that uses the costs of the high-level nodes as their $f$-values. The cost of a high-level node is the sum of the travel times along its paths. When CBS expands a high-level node $N$, it checks whether the node is a goal node. A high-level node is a goal node if and only if none of its paths collide. If $N$ is a goal node, then CBS terminates successfully and outputs the paths in $N$ as solution. (Thus, the fewer collisions there are in high-level nodes, the faster CBS terminates.) Otherwise, at least two paths collide. CBS chooses a collision to resolve and generates two high-level children of $N$, called $N_1$ and $N_2$. Both $N_1$ and $N_2$ inherit the constraints of $N$. If the chosen collision is a vertex collision $(j,k,s,t)$, then CBS adds the vertex constraint $(j,s,t)$ to $N_1$ (that prohibits agent $j$ from occupying vertex $s$ at time step $t$) and the vertex constraint $(k,s,t)$ to $N_2$. If the chosen collision is an edge collision $(j,k,s_1,s_2,t)$, then CBS adds the edge constraint $(j,s_1,s_2,t)$ to $N_1$ (that prohibits agent $j$ from moving from vertex $s_1$ to vertex $s_2$ between time steps $t$ and $t+1$) and the edge constraint $(k,s_2,s_1,t)$ to $N_2$. During the generation of high-level node $N$, CBS performs a low-level search for the agent $i$ affected by the added constraint. The low-level search for agent $i$ is a (best-first) A* search that ignores all other agents and finds a minimum-cost path from the start vertex of agent $i$ to its goal vertex that is both feasible and respects the constraints of $N$ that involve agent $i$.

ECBS($w$) \cite{BSSF:SOCS:14} is a $w$-suboptimal variant of CBS whose high-level and low-level searches are focal searches rather than best-first searches. A focal search, like A*, uses an \open list whose nodes $n$ are sorted in increasing order of their $f$-values $f(n) = g(n) + h(n)$, where $h(n)$ are the primary heuristic values. Unlike A*, a focal search with suboptimality factor $w$ also uses a \focal list of all nodes currently in the \open list whose $f$-values are no larger than $w$ times the currently smallest $f$-value in the \open list. The nodes in the \focal list are sorted in increasing order of their secondary heuristic values. A* expands a node in the \open list with the smallest $f$-value, but a focal search instead expands a node in the \focal list with the smallest secondary heuristic value. If the primary heuristic values are admissible, then a focal search is guaranteed to be $w$-suboptimal. The secondary heuristic can favor nodes in the \focal list that are close to the goal but without the requirement of having to be consistent (or admissible). This allows it to exploit the leeway afforded by $w$ that A* does not have available. The high-level and low-level searches of ECBS($w$) are focal searches. During the generation of a high-level node $N$, ECBS($w$) performs a low-level focal search with \open list, $\open^i(N)$, and \focal list, $\focal^i(N)$, for the agent $i$ affected by the added constraint. The high-level and low-level focal searches of ECBS($w$) use measures related to the number of collisions as secondary heuristic values.

CBS+HWY($w$) \cite{CUK:SOCS:15} is a different $w$-suboptimal variant of CBS that uses the highway heuristic in its low-level search. Given a set of directed edges in the graph, deemed as the highway, the highway heuristic values correspond to shortest path distances in the same graph but with modified weights. Specifically, the cost of each edge that is not a highway edge is inflated by a factor $w$. Using the highway heuristic values biases the low-level searches to find paths that use highway edges, likely reducing the number of head-on collisions. Finally, iECBS($w$) \cite{CUKXAK:IJCAI:2016} is a $w$-suboptimal variant of ECBS that uses the highway heuristic as the secondary heuristic for ordering its \focal list.

\section{RANDOMIZED MAPF SOLVERS}
In this section, we describe simple ways in which we can incorporate randomness into M*, inflated-M*($w$), ECBS($w$) and iECBS($w$).

In the M* framework, randomization can be incorporated in the neighbor generation process. Both Operator Decomposition M* (ODM*) and Enhanced Partial Expansion M* (EPEM*) \cite{W:DISS:15} construct the possible neighbors of a vertex in an agent-by-agent fashion. M* takes each partial neighbor specifying the action of the first $N-1$ agents and creates a separate copy of it. M* appends to this copy each of the possible actions of agent $N$ if it is in the collision set of the vertex being expanded. Otherwise, it generates a single copy of the partial neighbor by appending the action dictated by the individual policy. The neighbors are then inserted into the \open list by a deterministic process that orders them by their $g$-values. As a result, the relative orders of neighbors with equal $g$-values depends on the order that they appear in the neighbor list, and thus on the labeling of the agents. Randomizing the labeling of the agents effectively changes how tie-breaking is done.


In the CBS framework, randomization can be incorporated in multiple places. For example, each run can be performed using a random permutation of the agents' labels. Given such a permutation, the solver performs a low-level focal search for each agent in that order. Such a low-level focal search tries to avoid collisions with paths found for previous agents. This bears a significant effect not only on the paths found in the high-level root node but also on which collisions will be resolved as search proceeds.

Randomization can also be used for choosing which high-level node to \emph{expand} next. Expanding a high-level node in i/ECBS($w$) is the process of replanning for the constrained agent in it. The deterministic version of i/ECBS($w$) uses a \focal list in the high-level search tree that maintains all high-level nodes with total cost up to $w$ times the minimum. It then picks a high-level node for expansion from this list based on a deterministic secondary heuristic. This choice from the \focal list can now be randomized. However, since the secondary heuristic---that is, the total number of conflicts in the paths represented by this high-level node---provides useful information, the choice does not have to be uniformly at random.

Randomization can also be used for choosing which high-level nodes to \emph{generate} next. Generating a pair of high-level nodes in i/ECBS($w$) is the process of identifying a conflict and adding two successor high-level nodes that resolve it. The deterministic version of i/ECBS($w$) chooses the earliest conflict for generation of successors. Here too, randomization can be used to choose a conflict. Since choosing the earliest conflict works well in practice, the choice can be biased towards earlier conflicts and does not have to be done uniformly at random.

Moreover, randomization can also be used for choosing which low-level node to expand next. The deterministic version of i/ECBS($w$) uses a \focal list in the low-level search tree that maintains all low-level nodes with cost up to $w$ times the minimum. It then picks a low-level node for expansion from this list based on a deterministic secondary heuristic. The secondary heuristic value of a low-level node is the number of conflicts created by the partial path (from the start location to this node) with all other agents. Here too, randomization can be used to choose a node for expansion from the \focal list. Once again, this choice does not have to be done uniformly at random but can be biased according to the secondary heuristic.

Incorporating randomness in MAPF solvers has the additional benefit of parallelizability. For example, different random permutations of agents' labels, in both i/ECBS($w$) and M*, can be given to different spawned instantiations of the solvers to be run in parallel. The first instantiation that solves the problem instance terminates all other instantiations. This is both easy to implement, since different instantiations do not require communication between them, as well as likely to be beneficial when the runtimes exhibit a heavy-tailed distribution (as discussed in the next section).

\section{RRR STRATEGIES}
\begin{figure}
\centerfloat
  \begin{subfigure}[b]{0.24\textwidth}
    \centering
      	\includegraphics[width=1.1\textwidth]{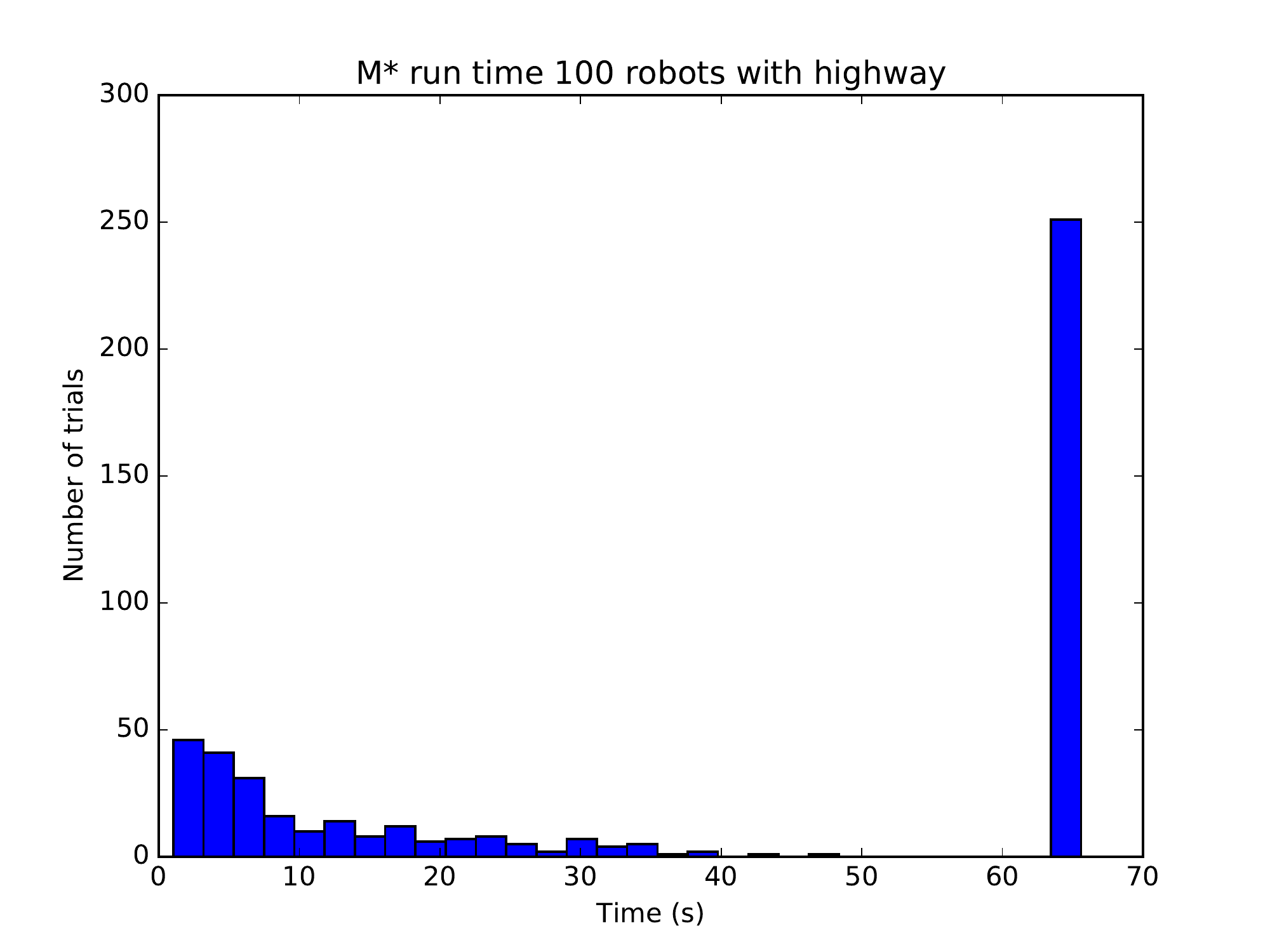}
    \caption{}
  \end{subfigure}
  ~
  \begin{subfigure}[b]{0.24\textwidth}
    \centering
      	\includegraphics[width=1.1\textwidth]{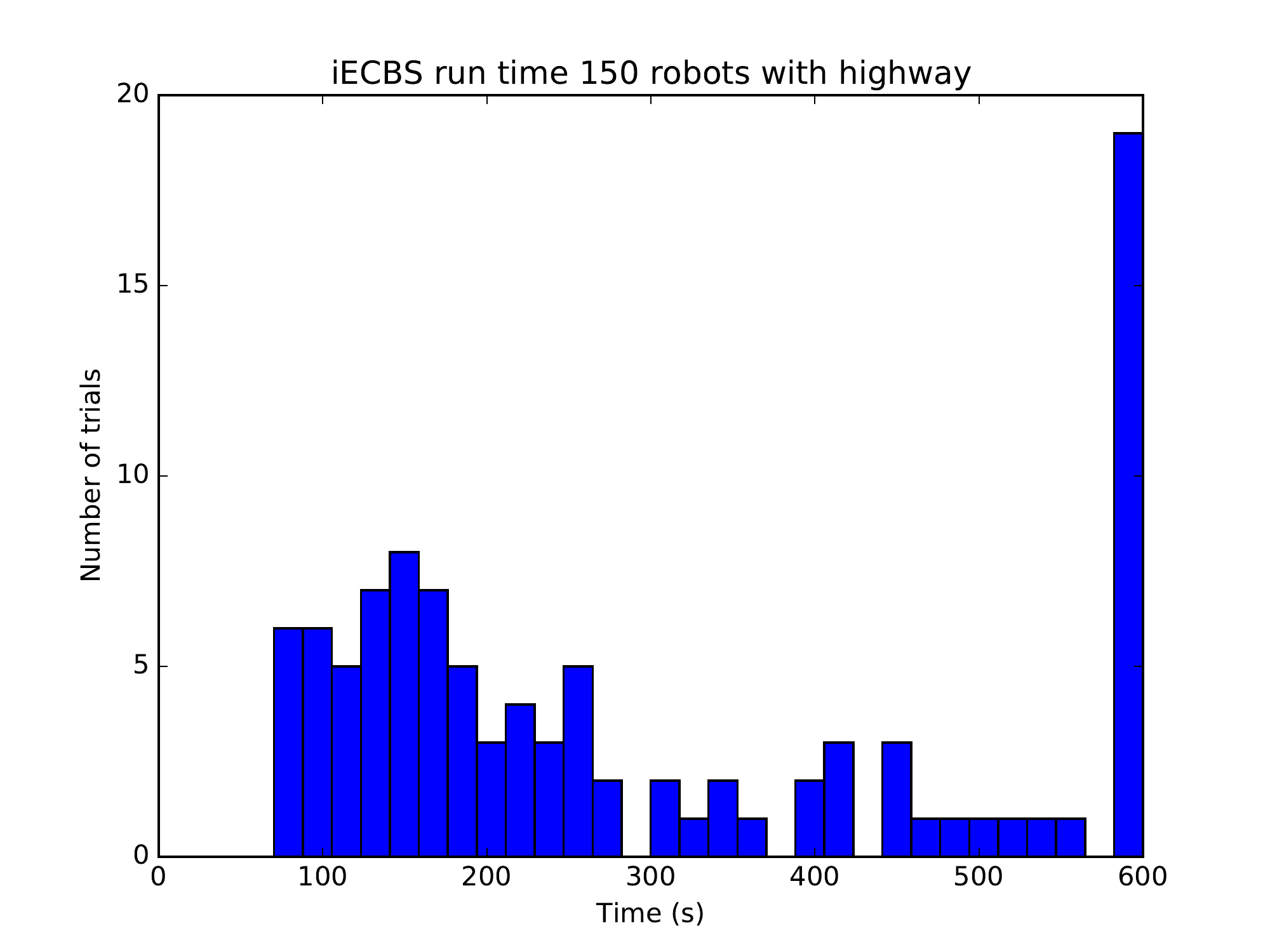}
    \caption{}
  \end{subfigure}

  \caption{Illustrates the heavy-tailed distribution of runtimes for M* and iECBS in (a) and (b), respectively. In (a), a fixed instance with $100$ agents (robots) is used with different runs of the randomized M* solver. In (b), a fixed instance with $150$ agents (robots) is used with different runs of the randomized iECBS solver.}
  \label{heavy_tailed_fig}
\end{figure}

The Normal distribution and many other probability distributions have exponentially decreasing tails. This means that outliers in such distributions are very rare and hence have lower impact on their mean, variance or higher moments. However, some distributions have tails that decay according to a power-law and are therefore deemed ``heavy-tailed''. The mean, variance or other higher moments of such heavy-tailed distributions may not converge. This is due to the large probability mass in their tails that allows outliers to have a greater impact on their moments.

Heavy-tailed distributions also occur in the runtimes of various search algorithms. They manifest themselves in the form of erratic means and variances. In such cases, the erratic mean and variance can be replaced by the more stable median and median absolute deviation (MAD). However, an erratic mean can also provide an inkling towards a heavy-tailed distribution that can possibly be exploited for computational benefits.

The occurrence of heavy-tails in the runtime distributions of search algorithms for various NP-hard combinatorial problems can be explained as follows. On the one hand, when a problem is NP-hard, it is unlikely to have an efficient deterministic algorithm for it that scales well with increasing problem size. From the theory of pseudo-randomness, it is strongly believed that a problem which can be solved using a polynomial-time randomized algorithm can also be solved using a polynomial-time deterministic algorithm by replacing the random bits used by the randomized algorithm with deterministically generated pseudo-random bits. Therefore, it is unlikely to have an efficient randomized algorithm that scales well for an NP-hard problem, establishing a negative theoretical result. On the other hand, AI research has focused on problem-specific heuristics that scale well in practice. Therefore, the only distribution of runtimes that is consistent with both (a) the long expected runtimes by virtue of the negative theoretical result, and (b) the frequent short runtimes in practice, is a heavy-tailed distribution.\footnote{Following the same reasoning, one can expect heavy-tailed distributions to be more apparent in the runtime behaviors of algorithms guided by stronger heuristics.}

For randomized algorithms that exhibit heavy-tailed behavior of runtimes, one can reduce the runtime variance by an RRR strategy. Such a strategy uses the observation that a short run of the algorithm is almost as likely to solve the problem instance as a longer run. Therefore, multiple short runs that use independent random bits in their executions are more likely to solve a problem instance than one long run of equivalent accumulated time. We note, however, that the short runs cannot be arbitrarily short since they need to accommodate a reasonable search effort.

In the context of MAPF, replacing deterministic MAPF solvers by their randomized versions, as described in the previous section, opens up the possibility for employing RRR strategies if the randomized versions exhibit heavy-tailed distributions of their runtimes. Figure \ref{heavy_tailed_fig} shows that such heavy-tailed distributions indeed characterize the runtime behaviors of our randomized versions of M* and iECBS.

The RRR strategies adapted to our randomized MAPF solvers are as follows. In the M* framework, the way in which ties are broken between search nodes with the same $g$-values has a significant effect on runtime. This is because one node may lead to a clean resolution of a potential conflict while the other may either fail to fully resolve the conflict, leading to more required computation, or in the worst case, cause the agents trying to resolve one conflict to potentially collide with another set of agents attempting to resolve a different conflict, thereby generating a single large conflict that is exponentially more expensive to resolve.

In the CBS framework, each run is performed using a random permutation of the initial order in which individual paths are planned for the agents. The low-level focal searches in ECBS($w$) and iECBS($w$) try to avoid collisions with previously computed paths for other agents. This has a significant effect not only on the paths found in the high-level root node but also on which collisions need to be resolved as the high-level search proceeds. Thus, it has a significant effect on the runtime, which our RRR strategy exploits for computational benefits.

\section{EXPERIMENTAL RESULTS}

\begin{figure}[t]
\centerfloat
  \begin{subfigure}[b]{0.5\textwidth}
    \centering
  	\includegraphics[width=1\textwidth]{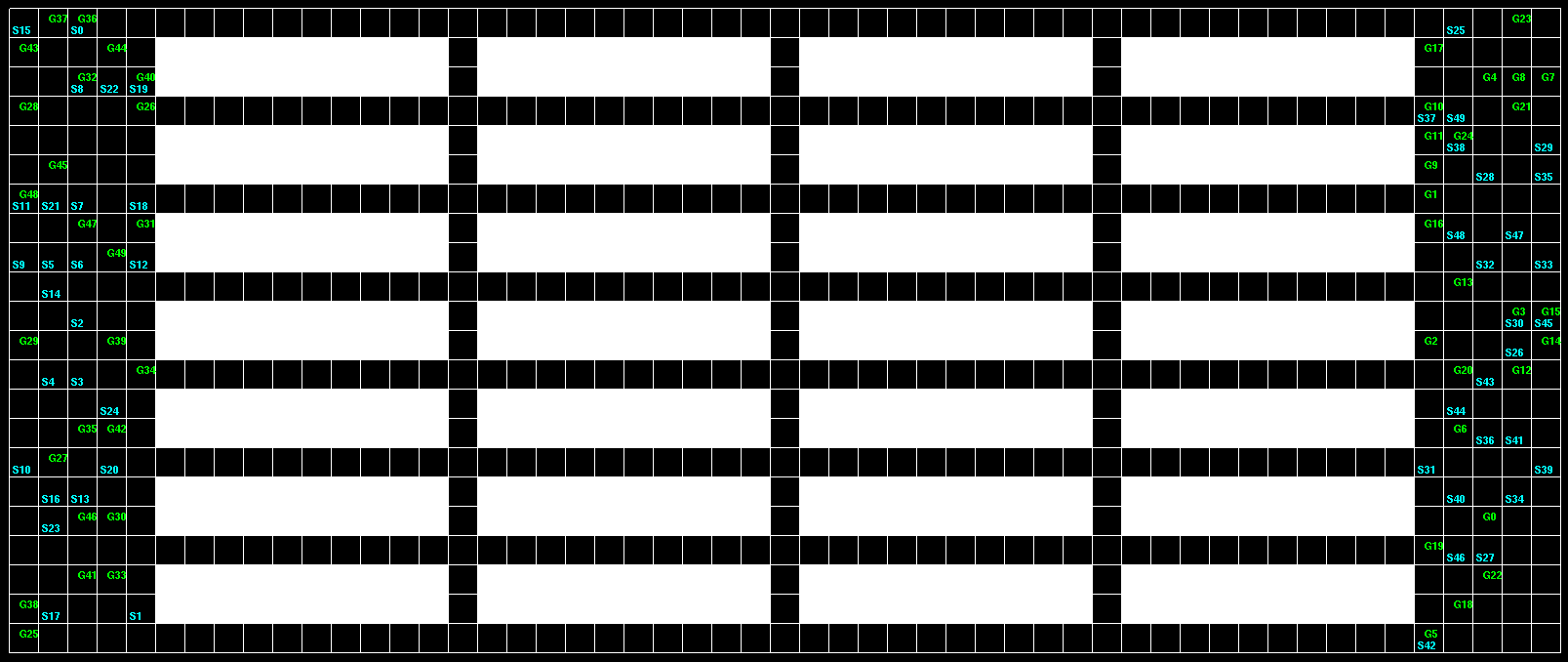}
    \caption{}
  \end{subfigure}
  
  \vspace{0.2cm}

  \begin{subfigure}[b]{0.5\textwidth}
    \centering
  	\includegraphics[width=1\textwidth]{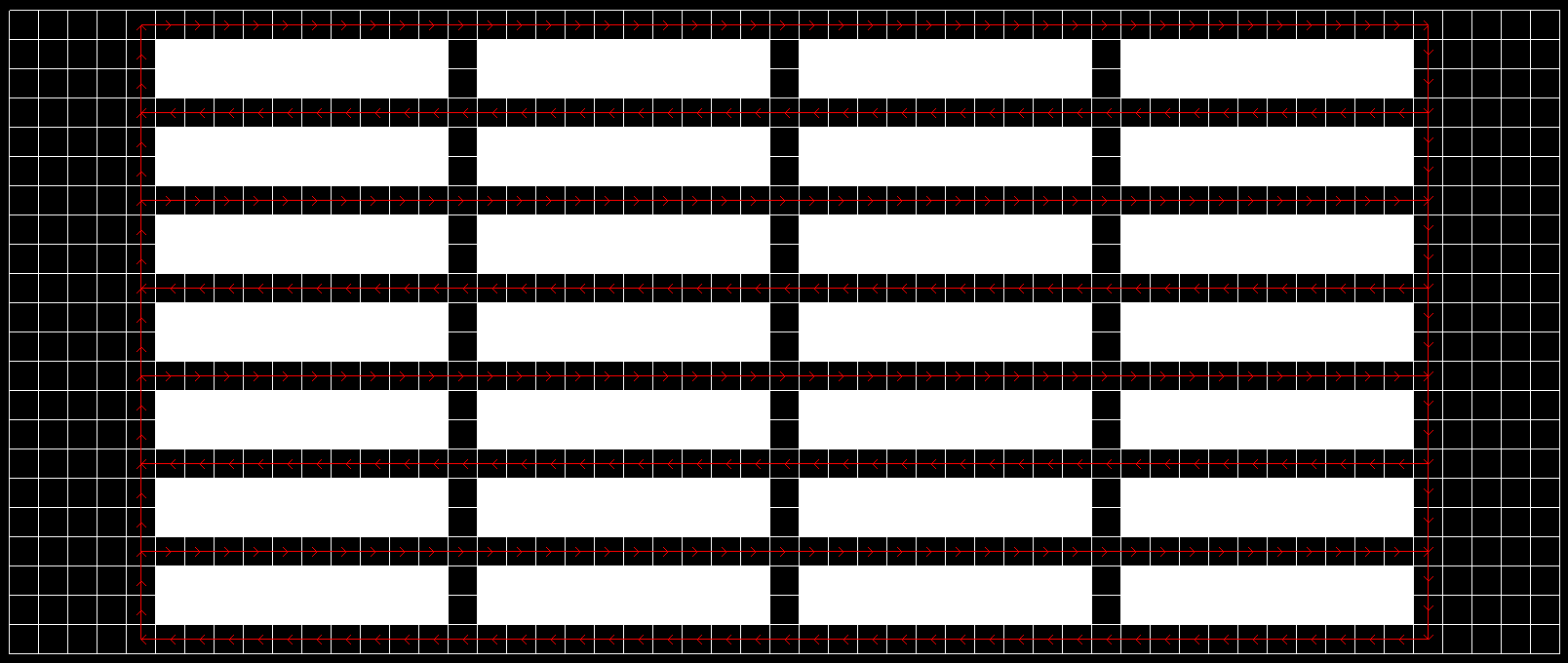}
    \caption{}
  \end{subfigure}

  \vspace{0.2cm}  
  
  \begin{subfigure}[b]{0.5\textwidth}
    \centering
  	\includegraphics[width=1\textwidth]{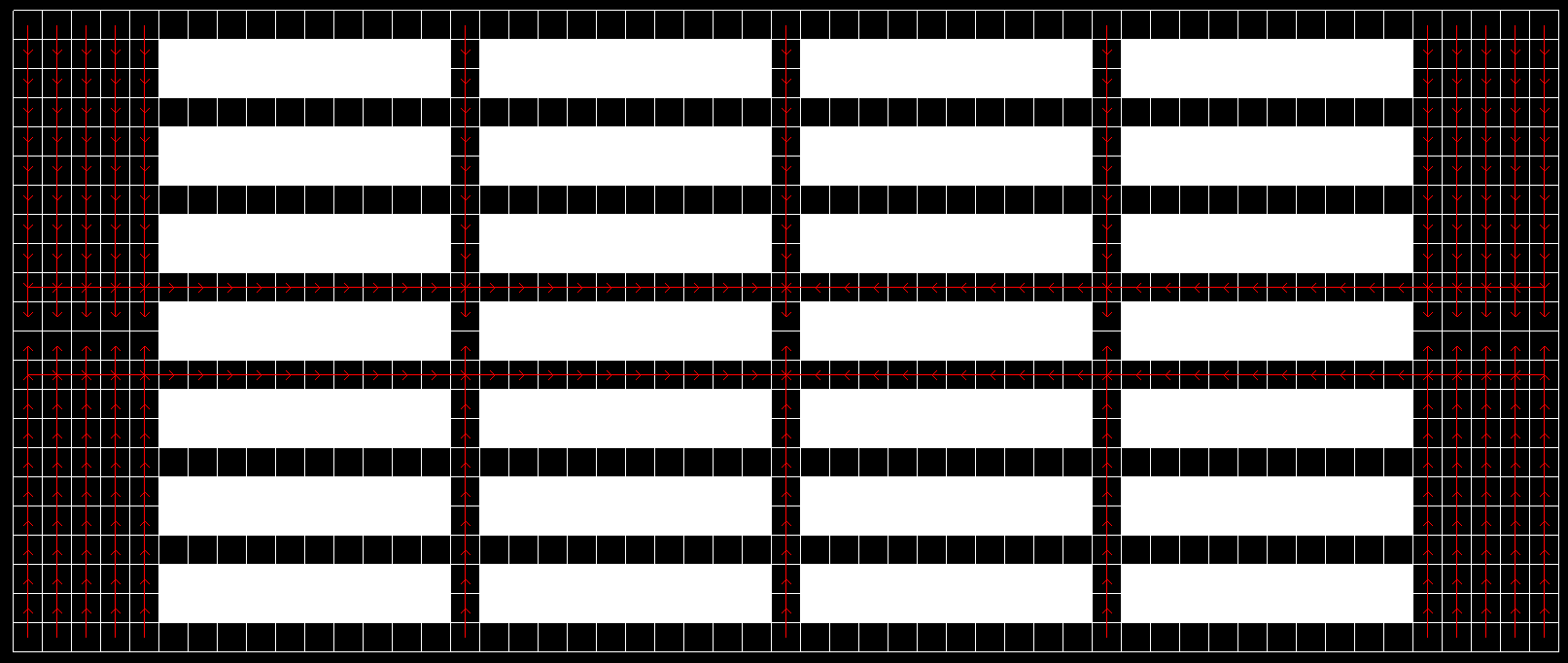}
    \caption{}
  \end{subfigure}

  \caption{(a) shows the Kiva-like map used in our experiments. (S)tart and (G)oal locations of $50$ agents are shown. (b) shows the ``positive'' highway used to effectively guide agents in our experiments. (c) shows the ``negative'' highway used to deliberately create congestion between agents.}
  \label{fig_hwys}
\end{figure}

In this section, we describe experiments performed to evaluate our RRR strategies. We show that these RRR strategies are beneficial in both the M* and CBS frameworks. The experiments were done on a cluster of $38$ Amazon EC2 c4.xlarge instances running on Intel(R) Xeon(R) CPU E5-2666 v3 @ 2.90GHz with 4 vcpu (2 physical cores) and 7.5GB RAM per instance. iECBS used $2$ workers per instance with each run using a $10$ minute timeout. M* used $1$ worker per instance with a $1$ minute timeout.\footnote{This is due to the high memory consumption of our M* implementation.}

The MAPF instances were chosen from the Kiva\footnote{now part of Amazon Robotics}-like domain as shown in Figure \ref{fig_hwys}(a). The Kiva-like domain is representative of warehousing applications of MAPF and is generally considered hard \cite{CUK:SOCS:15}. This is because significant coordination between the agents is required as many of them need to travel through narrow passageways to reach opposite sides of the map. We use $100$ randomly generated instances for each value of the number of agents (in increments of $10$). As illustrated in Figure \ref{fig_hwys}(a), given the number of agents, half of them are assigned a random start vertex in the left open space and a random goal vertex in the right open space, and vice-versa for the other half of the agents.

\begin{figure*}[!]
\centerfloat
 
  \begin{subfigure}[b]{0.33\textwidth}
    \centering
	\includegraphics[width=1.1\textwidth]{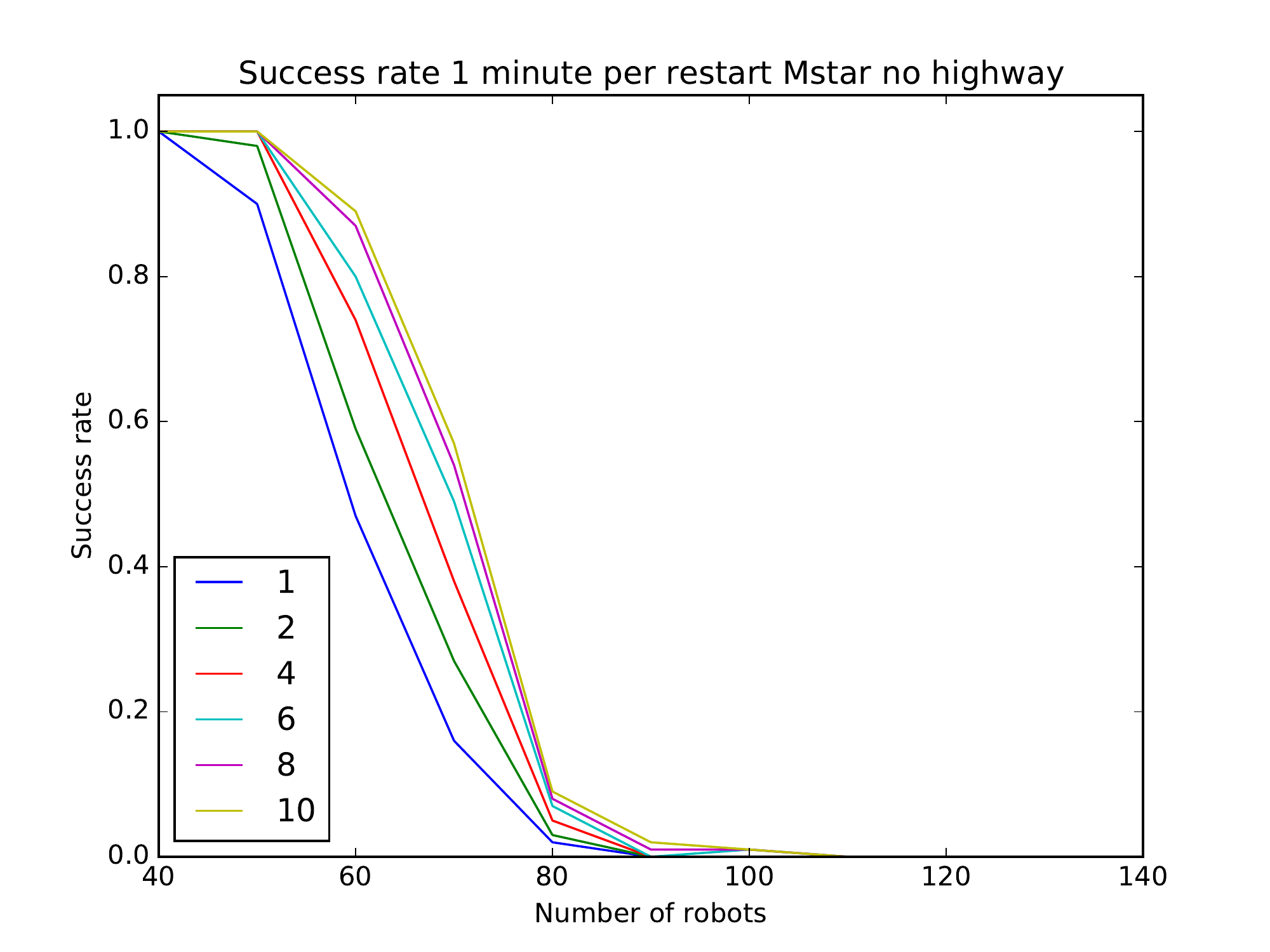}
    \caption{}
  \end{subfigure}
  ~
  \begin{subfigure}[b]{0.33\textwidth}
    \centering
	\includegraphics[width=1.1\textwidth]{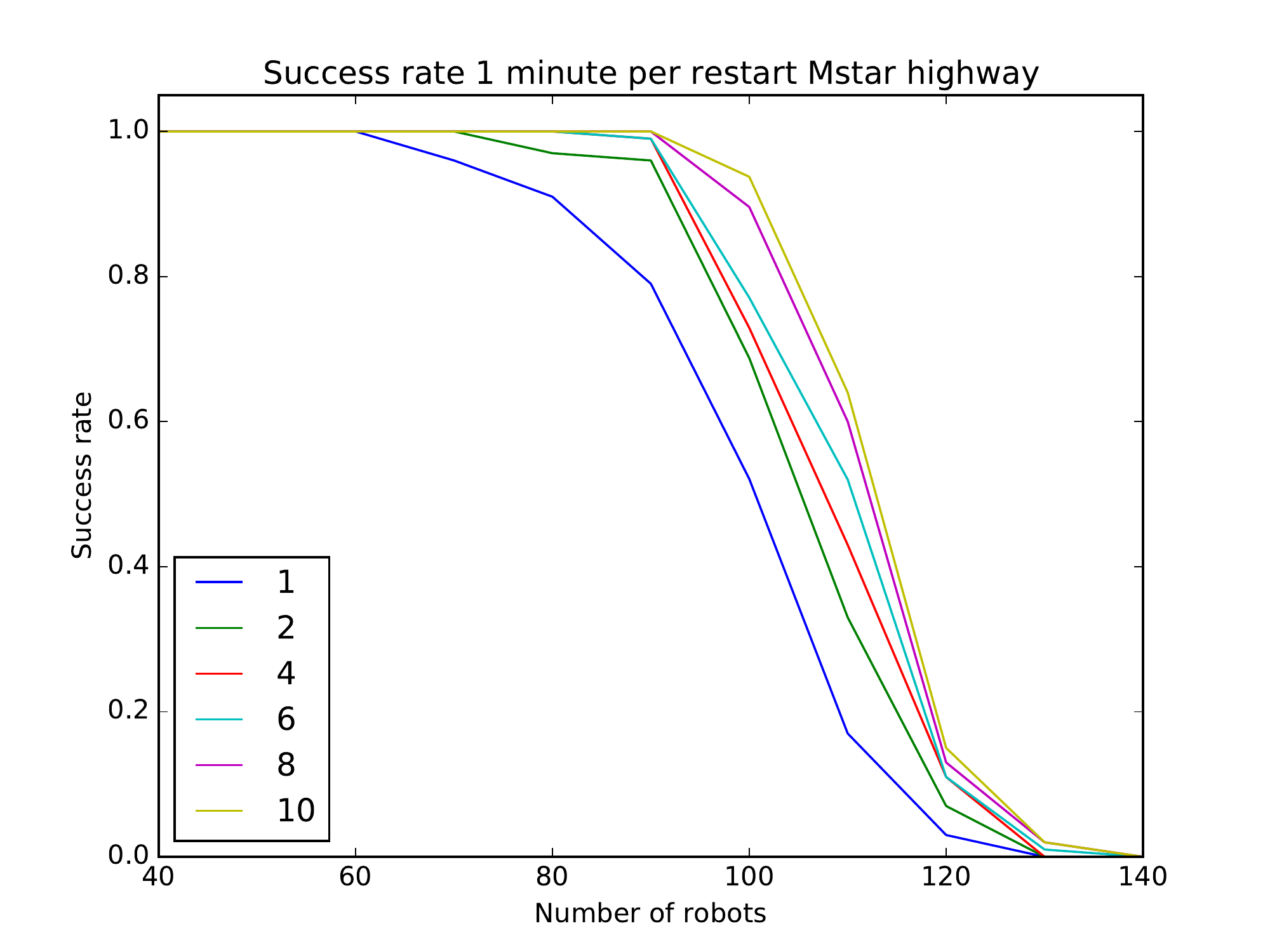}
    \caption{}
  \end{subfigure}
  ~
  \begin{subfigure}[b]{0.33\textwidth}
    \centering
	\includegraphics[width=1.1\textwidth]{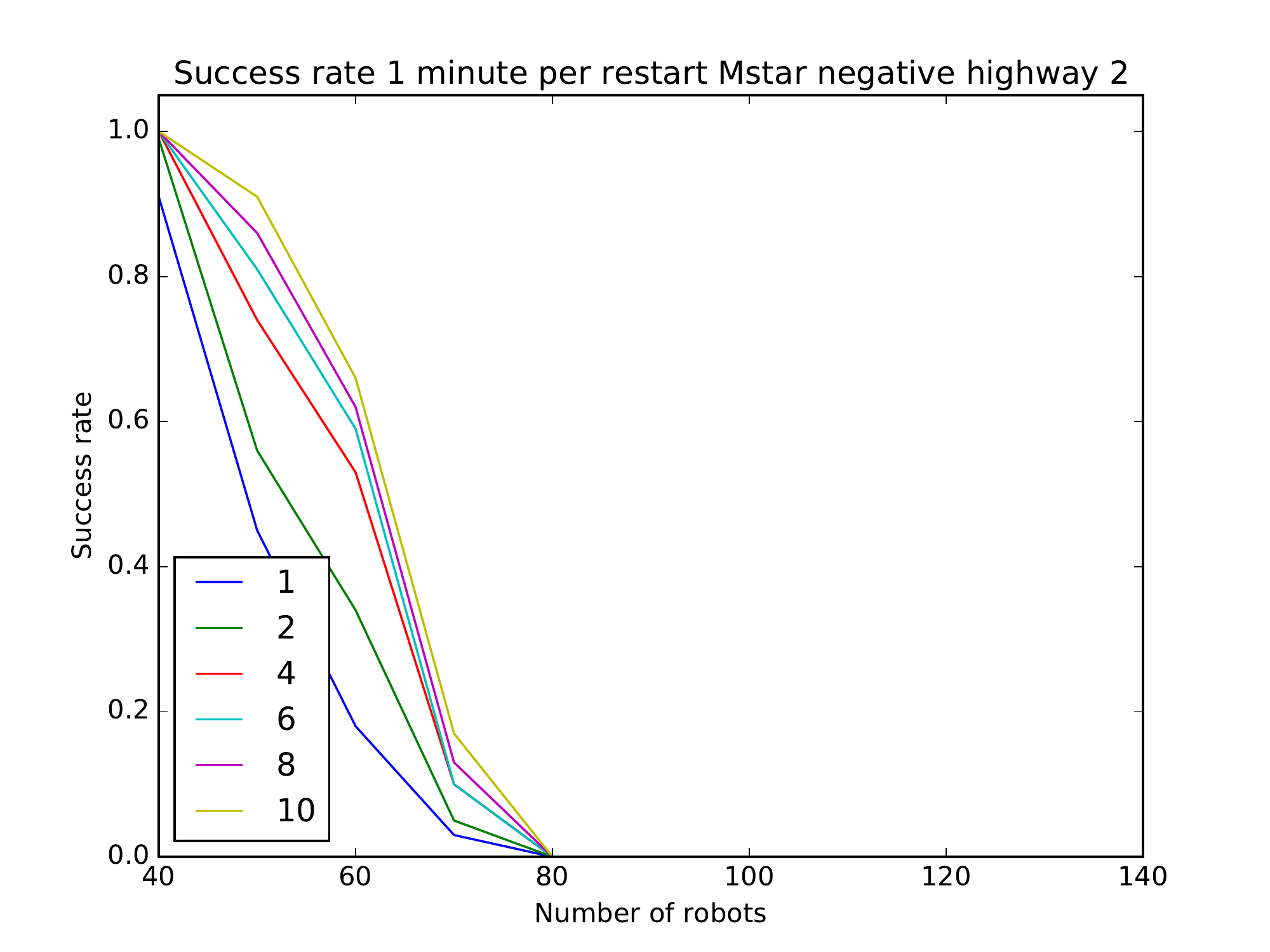}
    \caption{}
  \end{subfigure}
  
  \vspace{0.2cm}

  \begin{subfigure}[b]{0.33\textwidth}
    \centering
	\includegraphics[width=1.1\textwidth]{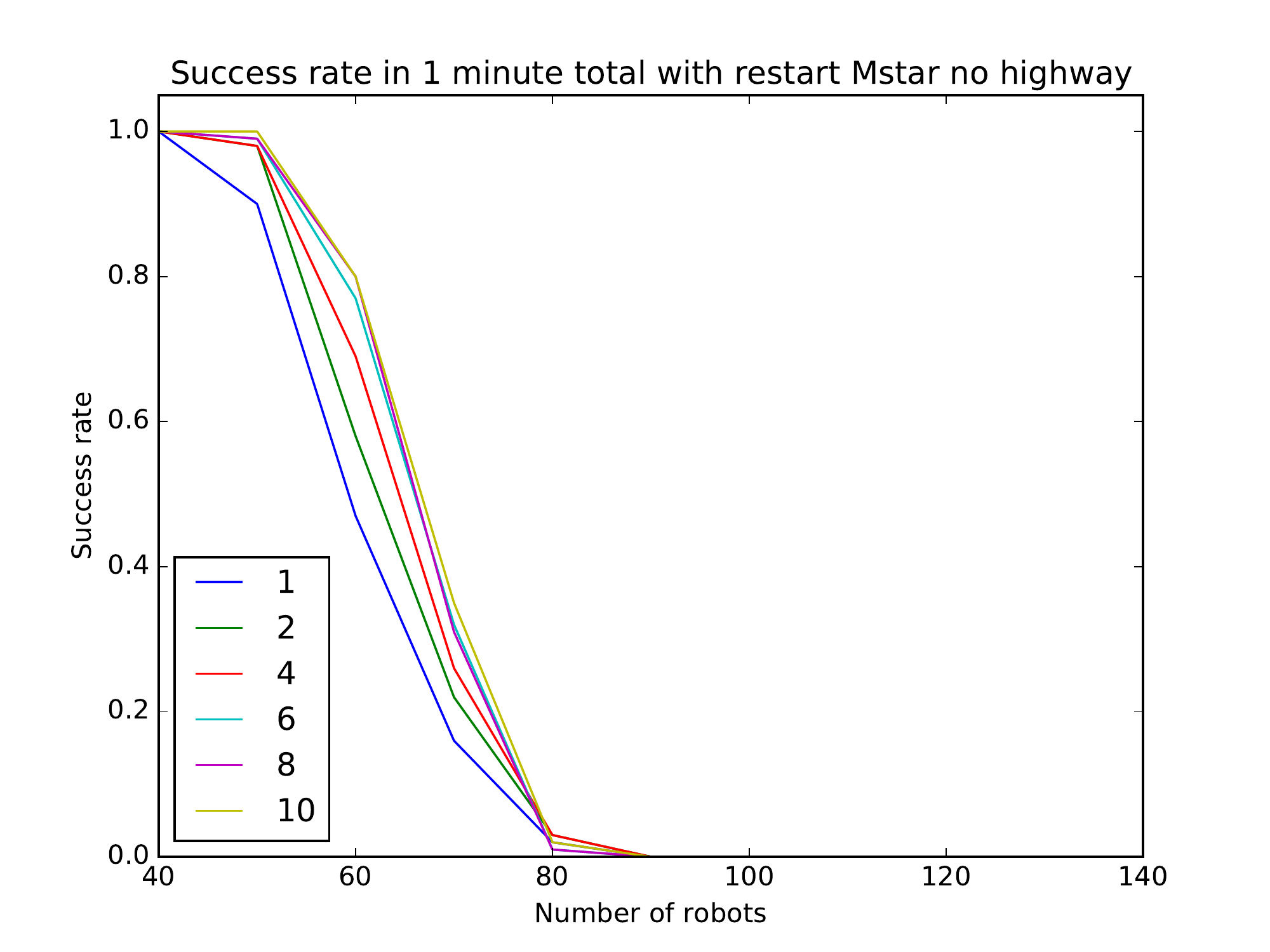}
    \caption{}
  \end{subfigure}
  ~
  \begin{subfigure}[b]{0.33\textwidth}
    \centering
	\includegraphics[width=1.1\textwidth]{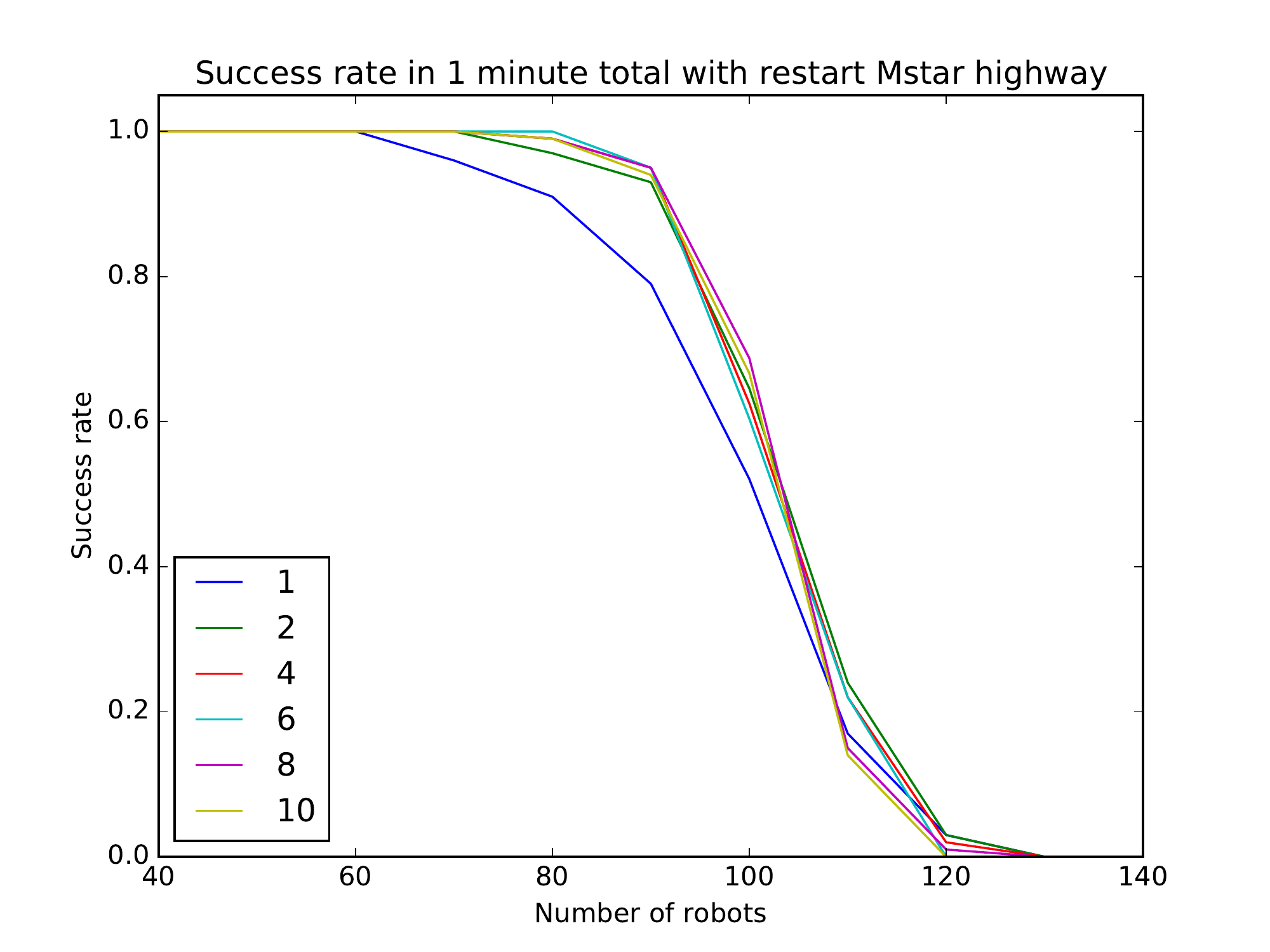}
    \caption{}
  \end{subfigure}
  ~
  \begin{subfigure}[b]{0.33\textwidth}
    \centering
	\includegraphics[width=1.1\textwidth]{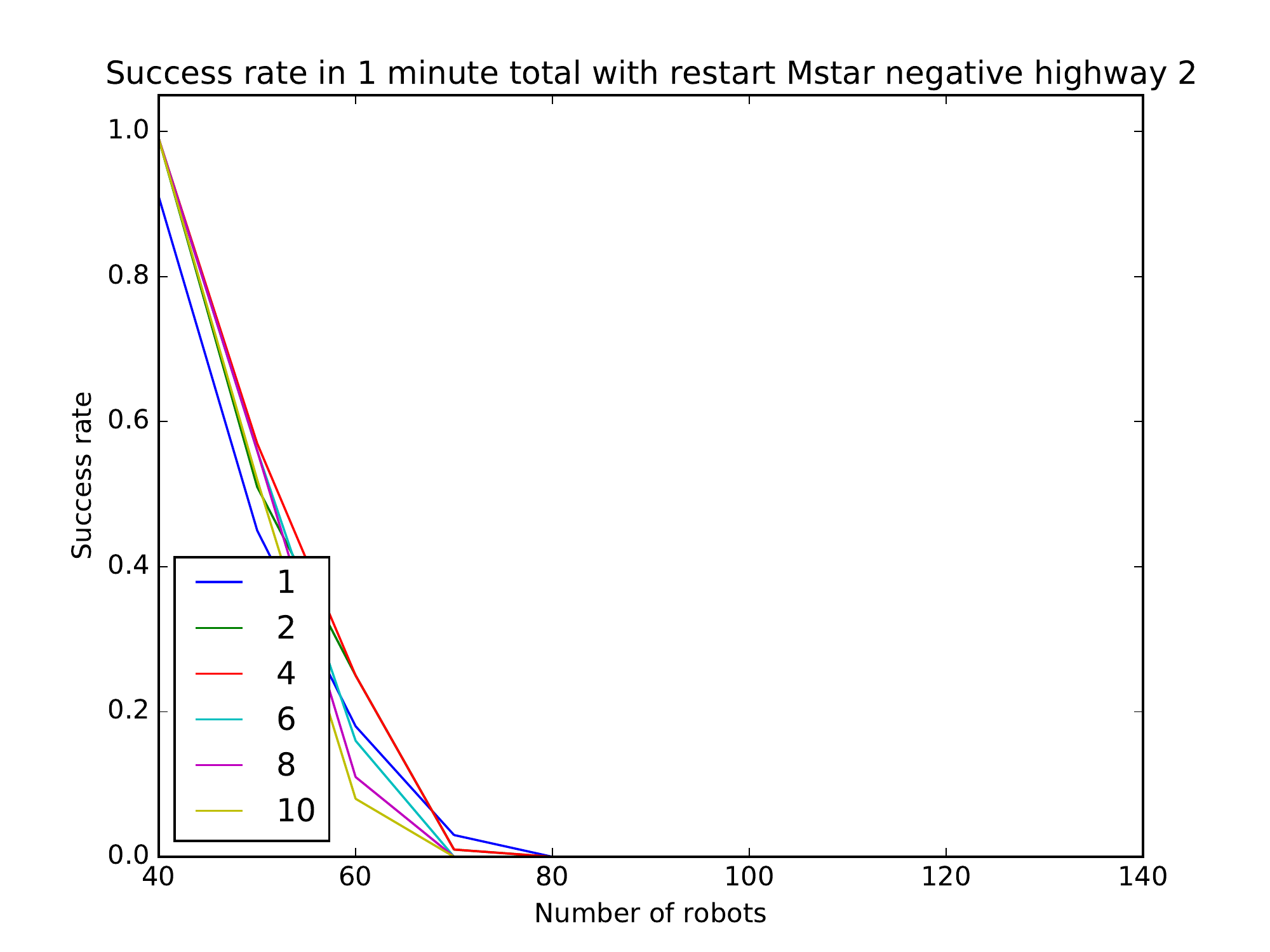}
    \caption{}
  \end{subfigure}
  
  \vspace{0.2cm}
  
  \begin{subfigure}[b]{0.33\textwidth}
    \centering
	\includegraphics[width=1.1\textwidth]{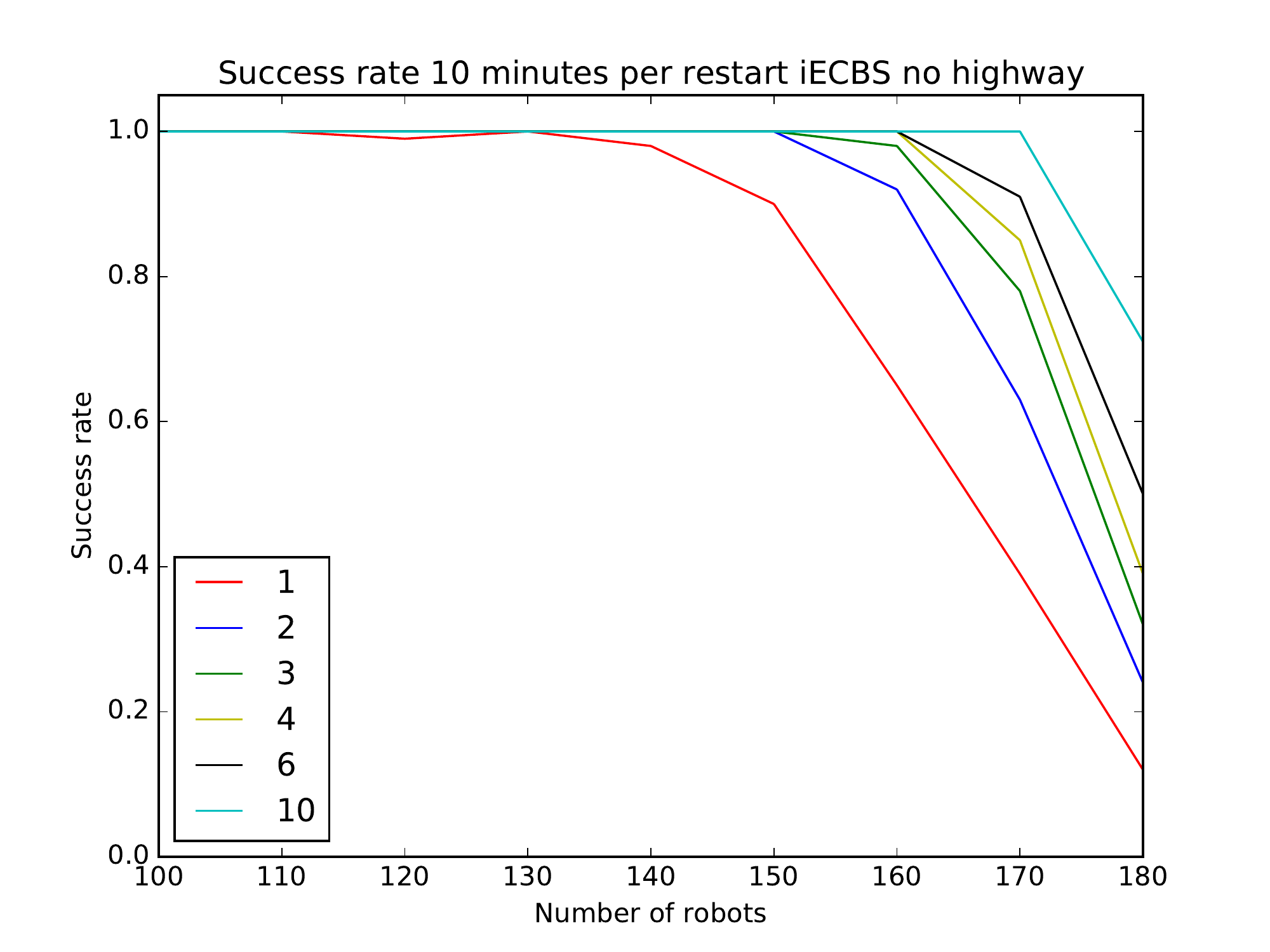}
    \caption{}
  \end{subfigure}
  ~
  \begin{subfigure}[b]{0.33\textwidth}
    \centering
	\includegraphics[width=1.1\textwidth]{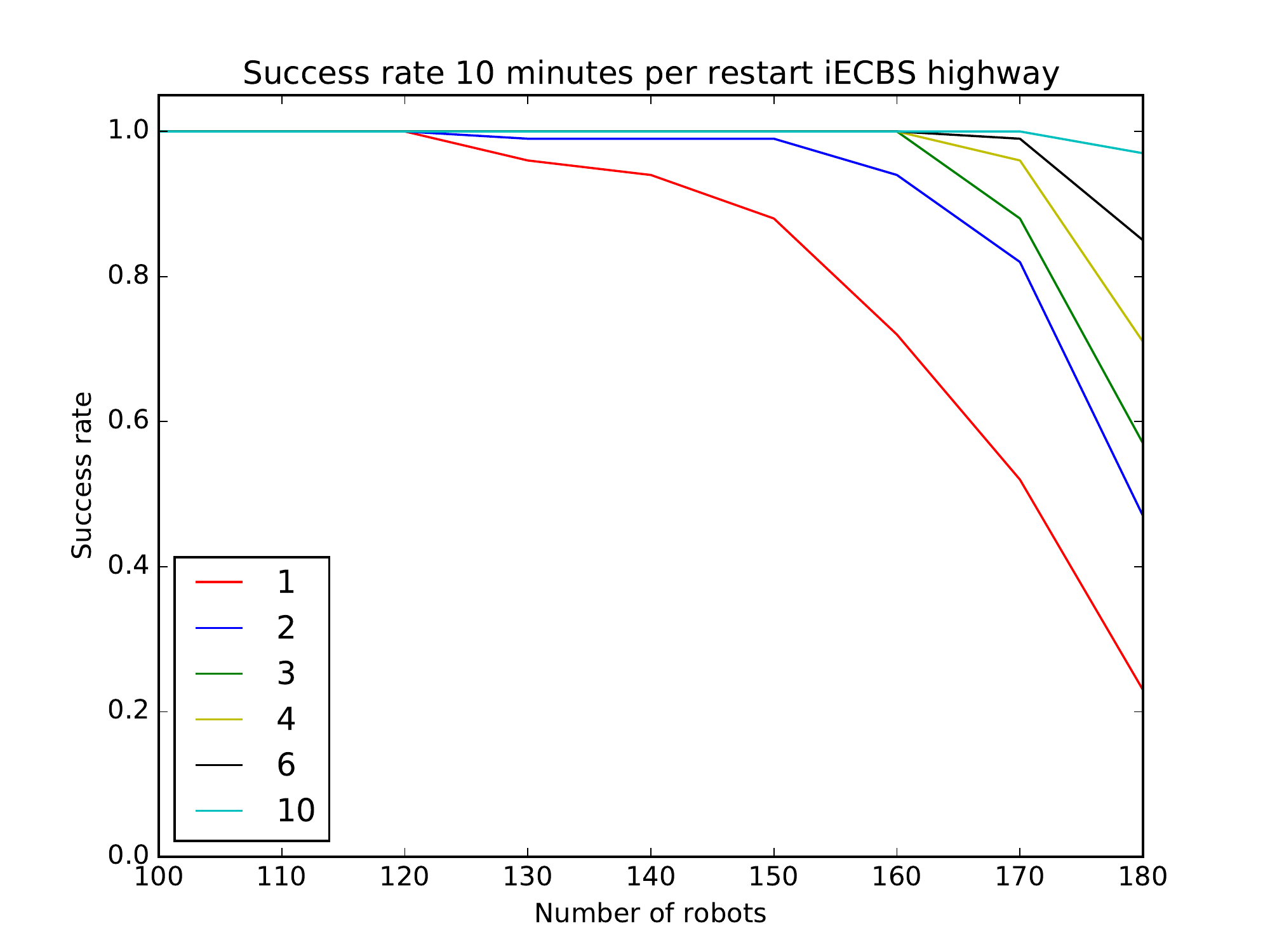}
    \caption{}
  \end{subfigure}
  ~
  \begin{subfigure}[b]{0.33\textwidth}
    \centering
	\includegraphics[width=1.1\textwidth]{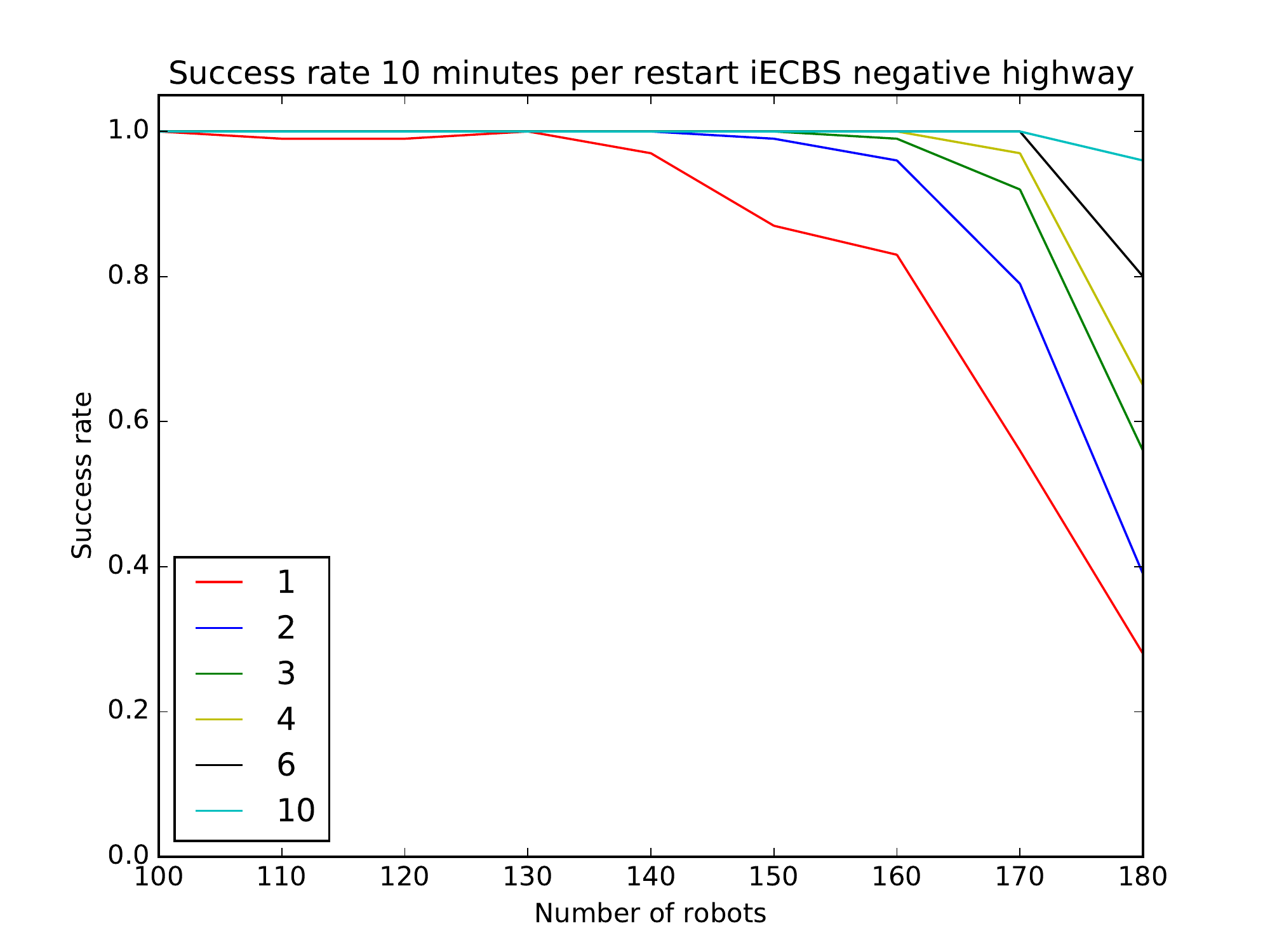}
    \caption{}
  \end{subfigure}
  
  \vspace{0.2cm}

  \begin{subfigure}[b]{0.33\textwidth}
    \centering
	\includegraphics[width=1.1\textwidth]{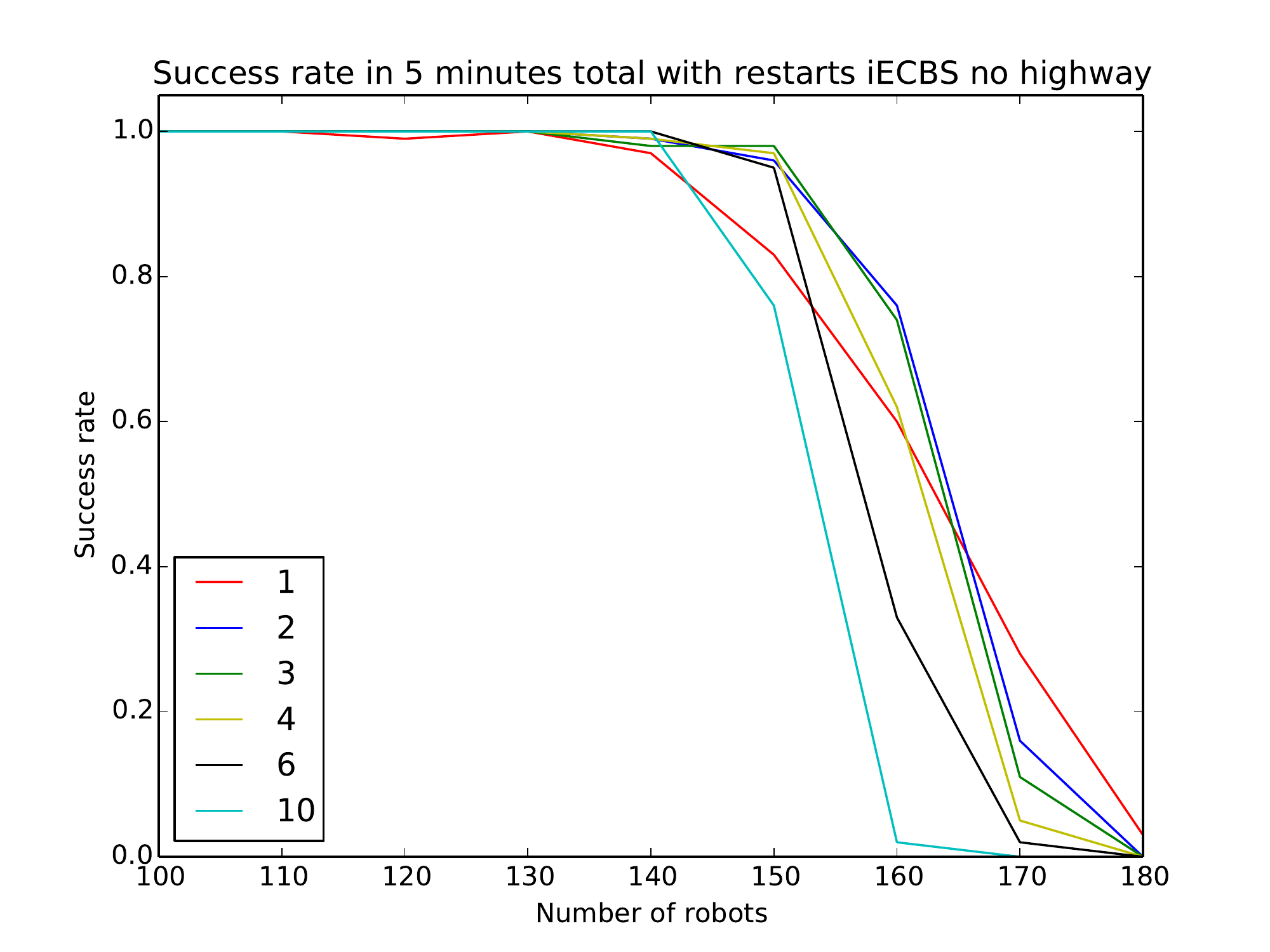}
    \caption{}
  \end{subfigure}
  ~
  \begin{subfigure}[b]{0.33\textwidth}
    \centering
	\includegraphics[width=1.1\textwidth]{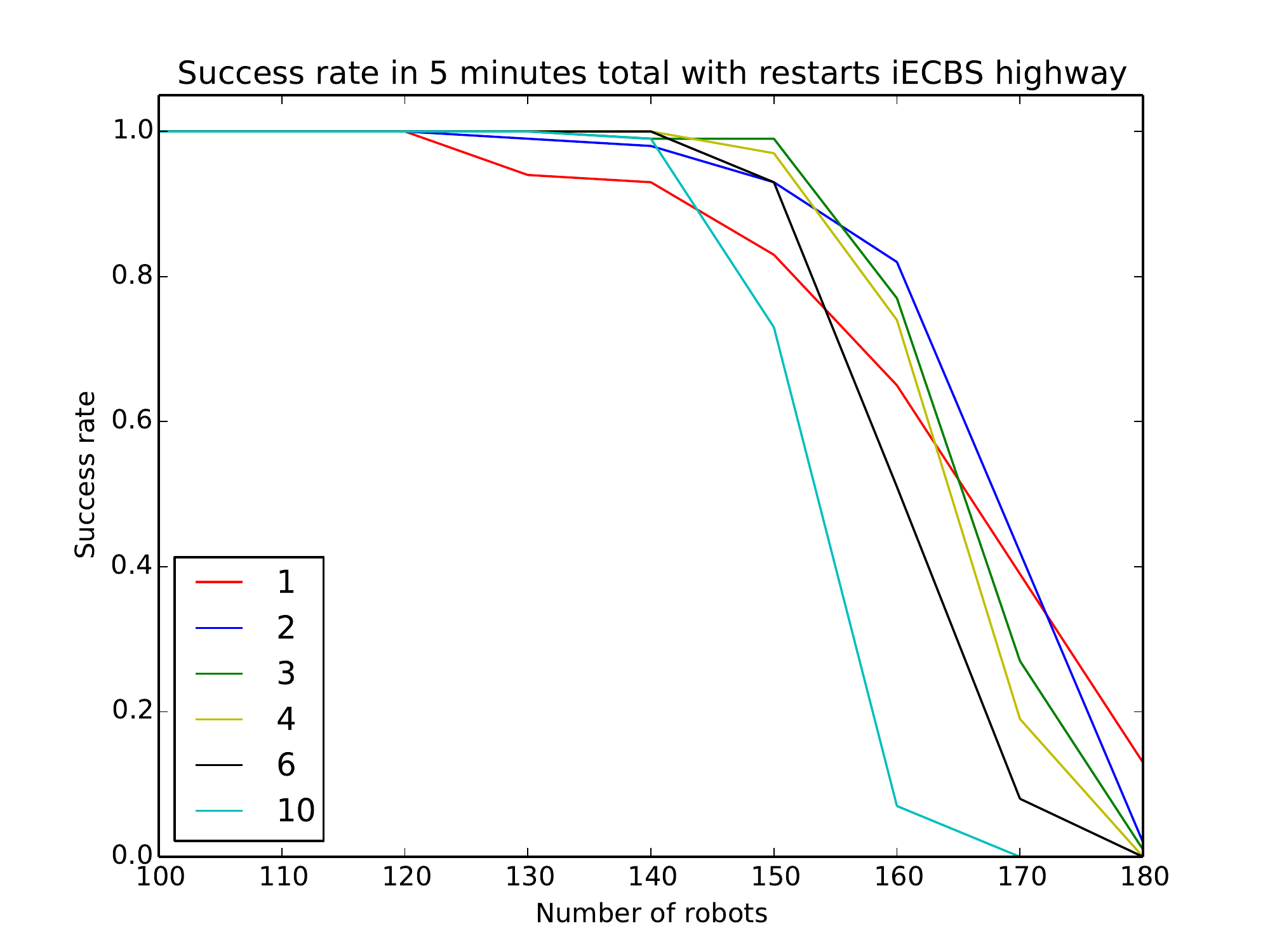}
    \caption{}
  \end{subfigure}
  ~
  \begin{subfigure}[b]{0.33\textwidth}
    \centering
	\includegraphics[width=1.1\textwidth]{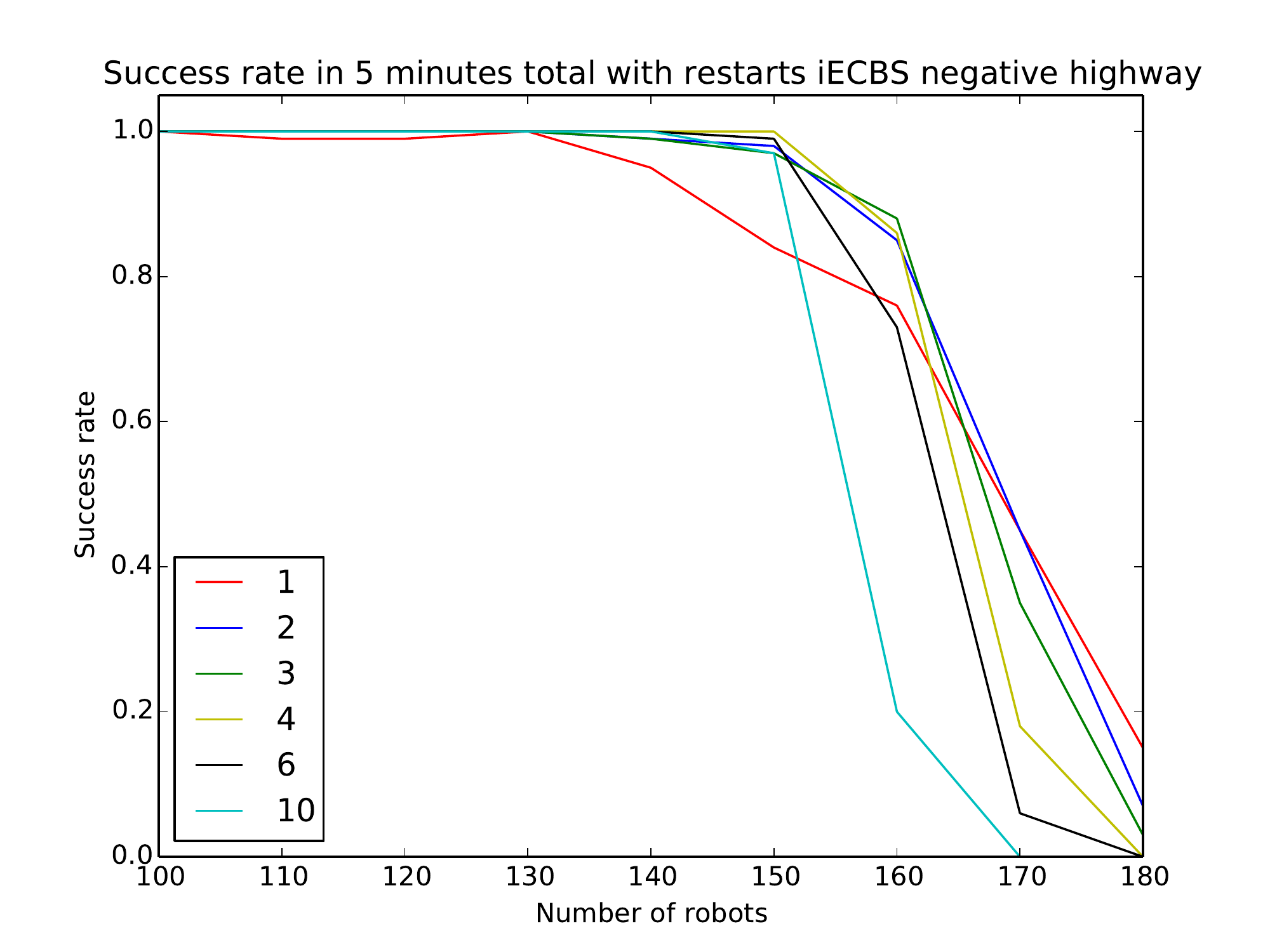}
    \caption{}
  \end{subfigure}
      
  \caption{Illustrates the benefits of RRR strategies in the M* and iECBS frameworks on Kiva-like instances. Both frameworks are tested under three categories: without highways, with a ``positive'' highway, and with a ``negative'' highway. Each colored curve corresponds to a different number of restarts. The x-axis shows the number of agents. The y-axis shows the percentage of instances solved within a certain time limit. For (a), (b) and (c), M*'s time limit is $1$ minute per trial. For (d), (e) and (f), M*'s time limit is $1$ minute divided by the number of restarts. For (g), (h) and (i), iECBS's time limit is $10$ minutes per trial. For (j), (k) and (l), iECBS's time limit is $10$ minutes divided by the number of restarts.}
  \label{fig_experiments}
\end{figure*}

Figure \ref{fig_experiments} shows the general trend in the difficulty of solving Kiva-like instances with increasing number of agents. The difficulty reflects as the success rate, i.e., the percentage of instances solved within a given time limit. A sharp decline in success rates is characteristic of randomly generated instances of NP-hard problems undergoing phase-transitions. Instances on the left of the phase transition do not require much coordination. This means that a solver can easily recover from bad decisions in the search process and likely solve them within the time limit. Instances on the right of the phase transition require a lot of coordination and hence are either not solvable or only a small fraction of them can be solved within the time limit. Instances in the phase transition require a critical amount of coordination and hence intelligent decisions need to be made by the solvers. These instances therefore serve as good test cases for distinguishing the strengths of different solvers.

Highways are directed edges that can be either human-generated or automatically generated \cite{CUKXAK:IJCAI:2016}. They are intended to exploit problem structure by incentivizing agents to construct paths that include their edges. This, in turn, encourages a global behavior of the agents that avoids collisions. Figure \ref{fig_hwys}(b) shows an example of a human-generated highway that encourages agents to traverse narrow passages in a single direction, thereby avoiding collisions. Hence, this highway is likely to be useful in improving runtime (and success rate) in our experiments. We refer to it as a ``positive'' highway. Note that any set of directed edges qualifies as a highway although it may not improve the runtime of solvers. Specifically, a set of edges that encourages congestion, for example, as shown in Figure \ref{fig_hwys}(c), is likely to degrade runtime (and success rate). We refer to this highway as a ``negative'' highway.

In our experiments, M* could run for a maximum of $1$ minute before exhausting available memory. The first row in Figure \ref{fig_experiments} simply reports on the success rate of M* with increasing number of trials. Here, each trial is given a time limit of $1$ minute. As expected, the success rate increases monotonically with the number of trials. The second row reports on the success rate of M* with our RRR strategy. Here, the total allocated runtime of $1$ minute is divided evenly among the trials. For example, the red line representing $4$ restarts gives each trial $15$ seconds.

Figures \ref{fig_experiments}(b) and (e) show how the use of the positive highway from Figure \ref{fig_hwys}(b) ``shifts'' the phase transition region to the right. This implies that instances involving more agents can now be solved under the same time limit compared to not using highways (Figures \ref{fig_experiments}(a) and (d)). Similarly, Figures \ref{fig_experiments}(c) and (f) show how the use of the negative highway from Figure \ref{fig_hwys}(c) shifts the phase transition region to the left. This implies that some instances that could previously be solved without using highways now become unsolvable under the same time limit.

We notice that our RRR strategy generally boosts the success rate of M* in all $3$ categories---that is, with no highway, with positive highway and with negative highway. However, we also notice that simply increasing the number of restarts does not always yield a higher success rate. This is because each trial should be given a reasonable amount of time to have a decent chance of finding a solution. For example, we notice that $10$ restarts is not optimal for all cases. In fact, when the search is expected to be harder, like in the case of the negative highway and increasing number of agents, time is better utilized in longer trials. This explains why, for $60$ agents in Figure \ref{fig_experiments}(f), $4$ restarts achieve higher success rate compared to $10$ restarts.


The third row in Figure \ref{fig_experiments} simply reports on the success rate of iECBS with increasing number of trials. Here, the problem instances have higher numbers of agents compared to the ones used for M*. Unlike M*, our implementation of iECBS is not memory-intensive and can therefore run for a longer time. Thus, we set the time limit of each trial to $10$ minutes. As expected, the success rate of iECBS increases monotonically with the number of trials. The fourth row reports on the success rate of iECBS with our RRR strategy. Here, the total allocated runtime of $10$ minutes is divided evenly among the trials.

Figures \ref{fig_experiments}(h), (i), (k) and (l) show that the use of positive or negative highways do not have a significant influence on the runtimes of iECBS. This is because the highways are used only in the secondary heuristic that guides the search. This secondary heuristic is used only to order nodes in the \focal list. In fact, even in the secondary heuristic, the highway heuristic values are used only to break ties among nodes with the same number of conflicts. More details of this machinery can be found in Figure 1 of \cite{CUKXAK:IJCAI:2016}.

Nevertheless, our RRR strategy boosts the success rate of iECBS in all $3$ categories. As in M*, here too, we notice that simply increasing the number of restarts does not always yield a higher success rate. This is because each trial should be given a reasonable amount of time to have a decent chance of finding a solution. In fact, the optimal number of restarts depends on the hardness of the problem instance characterized by the number of agents. For example, $2$, $3$ and $4$ restarts dominate the success rate for up to about $170$ agents. Beyond this point, the problem instances are even harder and require longer runs.


\section{CONCLUSIONS}

MAPF is a well-studied problem in AI and robotics with many real-world applications. In this paper, we recognized the deterministic nature of existing state-of-the-art MAPF solvers, like M* and iECBS, which leads to their poor performance on hard instances. We developed randomized versions of these solvers to study their amenability to RRR strategies. The randomized versions of these solvers replaced arbitrary choices with random ones in the search processes. On a given instance, the runtimes of these solvers exhibit heavy-tailed distributions that were exploited using RRR strategies with the intuition that, given a hard instance, multiple short runs have a better chance of solving it compared to one long run. We validated this intuition through experiments using a Kiva-like domain and showed that our RRR strategies increased the success rates of M* and iECBS. In future work, we intend to use randomness in additional ways and validate them.

\bibliography{LirBib}
\bibliographystyle{aaai}

\end{document}